\pdfoutput=1
\documentclass[11pt]{article}
\usepackage[final]{acl}
\usepackage{times}
\usepackage{latexsym}
\usepackage[T1]{fontenc}
\usepackage[utf8]{inputenc}
\usepackage{microtype}
\usepackage{inconsolata}
\usepackage{graphicx}
\usepackage{amsmath}
\usepackage{enumitem}
\usepackage{cleveref}
\crefname{figure}{Fig.}{Figs.}
\crefname{table}{Tab.}{Tabs.}
\crefname{section}{Sec.}{Secs.}
\crefname{subsection}{Subsec.}{Subsecs.}
\crefname{equation}{Eq.}{Eqs.}
\crefname{theorem}{Thm.}{Thms.}
\crefname{lemma}{Lem.}{Lems.}
\crefname{algorithm}{Alg.}{Algs.}
\usepackage{amssymb}
\usepackage{pifont}
\newcommand{\circlednum}[1]{\ding{\numexpr171+#1\relax}}
\usepackage{url}
\usepackage{hyperref}

\title{Representation Potentials of Foundation Models for \\ Multimodal Alignment: A Survey}

\author{Jianglin~Lu$^{1}$\thanks{Corresponding author: \texttt{jianglinlu@outlook.com}.}, Hailing~Wang$^{1}$, Yi~Xu$^{1}$, Yizhou~Wang$^{1}$, Kuo~Yang$^{1}$, Yun Fu$^{1,2}$ \\
  $^{1}$Department of Electrical and Computer Engineering, Northeastern University \\
  $^{2}$Khoury College of Computer Science, Northeastern University\\
  Resouce: \href{https://github.com/Jianglin954/Representation-Alignment-Survey}{https://github.com/Jianglin954/Representation-Alignment-Survey}
}

\begin{document}
\maketitle

\begin{abstract}
Foundation models learn highly transferable representations through large-scale pretraining on diverse data. 
An increasing body of research indicates that these representations exhibit a remarkable degree of similarity across architectures and modalities.
In this survey, we investigate the \textit{representation potentials} of foundation models, defined as the latent capacity of their learned representations to capture task-specific information within a single modality while also providing a transferable basis for alignment and unification across modalities.
We begin by reviewing representative foundation models and the key metrics that make alignment measurable.
We then synthesize empirical evidence of representation potentials from studies in vision, language, speech, multimodality, and neuroscience.
The evidence suggests that foundation models often exhibit structural regularities and semantic consistencies in their representation spaces, positioning them as strong candidates for cross-modal transfer and alignment. 
We further analyze the key factors that foster representation potentials, discuss open questions, and highlight potential challenges.

\end{abstract}
\section{Introduction}

Foundation models, trained through large-scale pretraining on vast and heterogeneous data, have driven remarkable progress and significantly accelerated the pursuit of artificial general intelligence \cite{bommasani2021opportunities, cui2022can, firoozi2023foundation, azad2023foundational, zhou2024comprehensive}.
By acquiring highly transferable and general-purpose representations, they have become the backbone of a wide spectrum of applications, spanning natural language processing \cite{roberta, deberta, rajendran2024causal}, computer vision \cite{ViTs, convnet, ConvNeXtV2, dinov3}, speech processing \cite{belinkov2017, wav2vec2, whisper}, robotics \cite{brohan2022rt, team2025gemini}, and medical domains \cite{moor2023foundation, huang2024enhancing, khan2025comprehensive}.

A growing body of research has shown that the representations learned by foundation models are not only powerful in isolation but also exhibit strong similarity across architectures, training objectives, and even modalities \cite{alignmentforum_natural_abstraction, ng2023can, llava, visioncheckup, Platonic, Maniparambil, wang2025towards}. 
We refer to this capacity as the \textit{representation potential} of foundation models. 
This perspective carries significant implications: if foundation models naturally converge toward shared representational structures, they may approximate modality-agnostic abstractions and encode common statistical regularities of the world, even without explicit alignment.
Understanding these potentials is essential, not only for advancing scientific theories of representation learning but also for enabling practical benefits such as model interoperability, transferability, interpretability, and alignment with biological cognition.

In this survey, we focus on the representation potentials of unimodal foundation models, with the goal of assessing their capacity for alignment. We structure our discussion around four central themes. 
First, we introduce representative foundation models in vision, language, speech, and multimodality. Second, we review the metrics that make representation alignment measurable, including centered kernel alignment \cite{kornblith2019similarity}, canonical correlation analysis \cite{morcos2018insights}, and mutual nearest neighbors \cite{haghverdi2018batch}.
Third, we synthesize empirical evidence for representation potentials, drawing from studies in vision, language, speech, cross-modal alignment, and neuroscience.
Fourth, we analyze the key factors that drive representation potentials, such as scale, architectural inductive biases, training objectives, and task and instruction diversity.
Alongside these advances, we highlight pressing open questions: the limits of convergence across modalities, the need for robust evaluation standards, the influence of bias and sociotechnical context, and cases where domain-specific divergence may arise.

The remainder of this survey is organized as follows. In Section~\ref{preliminaries}, we introduce foundation models across modalities. Section~\ref{metrics} reviews major metrics for quantifying representation similarity and alignment. Section~\ref{convergence} presents evidence for representation potentials in vision, language, speech, cross-modal, and neuroscience contexts. Section~\ref{factors} analyzes the underlying drivers of alignment, including scale, architectures, training paradigms, and tasks. Section~\ref{sec7} discusses open questions and challenges. In Section~\ref{conclusion}, we conclude with key insights and directions for future research.
\section{Foundation Models}
\label{preliminaries}
This section provides a general definition of foundation models and then presents representative examples across computer vision, natural language processing, speech and multimodal domains.

\subsection{Definition}
\citet{bommasani2021opportunities} first introduced the term \textit{foundation model} to describe machine learning models trained on vast and diverse datasets, typically with large-scale self-supervision, that can be applied to a broad range of downstream tasks.
Three features distinguish foundation models from their earlier predecessors:
\circlednum{1} \textit{Broad data}: They are trained on extensive and diverse datasets, often collected at web scale, which provide robust and transferable representations.
\circlednum{2} \textit{Self-supervision}: They learn directly from raw, unlabeled data by predicting missing information or inherent structures, thus avoiding the reliance on large volumes of manually annotated data.
\circlednum{3} \textit{Adaptability}: Once trained, they can be fine-tuned, prompted, or otherwise adapted to a wide range of downstream tasks, underscores their general-purpose nature and their ability to serve as a foundation for numerous specialized applications.
Based on these characteristics, the following subsections briefly introduce representative foundation models in computer vision, language, speech, and multimodal learning.

\subsection{Vision Foundation Models}
Vision foundation models (VFMs) \cite{convnet, dinov3} are large-scale neural architectures designed to learn robust visual representations that transfer across tasks.
Canonical examples include ResNet \cite{he2016deep}, Vision Transformer (ViT) \cite{ViTs}, ConvNeXt \cite{ConvNeXtV2}, Dinov2 \cite{dinov2}, and the Segment Anything Model (SAM) \cite{kirillov2023segment}.
VFMs are typically trained on billion-scale image datasets using self-supervised learning, weakly supervised signals, or multimodal objectives.
Earlier vision models depended on task-specific annotated datasets, but VFMs provide universal feature embeddings that can be reused or lightly adapted.
These features support a wide range of applications, including image classification, object detection, and segmentation, as well as higher-level reasoning tasks such as visual question answering and captioning.
Recently, VFMs have also become essential in vision systems such as segmentation-anything frameworks \cite{kirillov2023segment, ravi2025sam}, image generative pipelines \cite{yang2023diffusion, zhang2023a}, and world models \cite{ha2018world, zhou2025dinowm}. This transition marks a shift in computer vision from narrowly specialized solutions to foundational infrastructures.

\subsection{Large Language Models}

Large language models (LLMs) \cite{gpt2, touvron2023llama, gpt4} are trained on massive text corpora to acquire broad linguistic and semantic knowledge.
Representative examples include BERT \cite{bert}, T5 \cite{T5}, Qwen \cite{qwen}, and the LLaMA-series \cite{llama3}, and conversational agents such as ChatGPT \cite{GPT3}.
By predicting masked spans or next tokens, LLMs capture both syntactic structures and semantic relations that generalize across diverse tasks. 
They can be adapted through fine-tuning, prompting, or in-context learning to applications such as summarization, translation, question answering, reasoning, and dialogue.
A defining feature is their scale: empirical studies show that performance improves predictably as the number of parameters, the volume of training data, and the compute budget increase \cite{kaplan2020scaling}. 
Beyond higher accuracy, larger models also exhibit emergent abilities that are absent in smaller counterparts \cite{wei2022emergent}.
These findings have shaped both research practices and industrial applications, positioning LLMs as the foundation for multimodal \cite{yin2024survey} and agentic extensions \cite{li2025review}.

\subsection{Speech Foundation Models}

Speech foundation models (SFMs) are trained on extensive audio corpora, focusing on speech signals while learning both acoustic-level and linguistic-level abstractions. 
Representative examples include wav2vec \cite{wav2vec, wav2vec2}, HuBERT \cite{hsu2021hubert}, WavLM \cite{chen2022wavlm}, Whisper \cite{whisper}, and SeamlessM4T \cite{barrault2023seamlessm4t}.
These models learn by predicting masked or latent units from raw waveforms, enabling them to serve as universal encoders of speech. 
SFMs support a wide variety of tasks, ranging from automatic speech recognition and speaker verification to emotion recognition, speech translation, and speech-to-speech generation \cite{cui2024recent, whisper}.
Beyond supervised fine-tuning, SFMs also demonstrate robust performance in zero-shot and few-shot settings. As a result, SFMs mark a decisive transition from specialized pipelines to broadly adaptable foundation-level architectures.

\subsection{Multimodal Foundation Models}
Multimodal foundation models (MFMs) integrate signals from multiple modalities, such as vision, language, audio, and video, into a unified architecture.
Early MFMs include CLIP \cite{clip} and ALIGN \cite{align}, which use image–text contrastive learning. Later developments such as BLIP \cite{blip} and CoCa \cite{coca} focus on vision–language generation, while Flamingo \cite{Flamingo} and PaLI \cite{chen2022pali} advance few-shot multimodal reasoning. More recent large-scale systems such as GPT-4 \cite{gpt4} and Gemini \cite{team2023gemini} demonstrate general-purpose multimodal intelligence.
Training MFMs typically combines contrastive alignment (for paired modalities such as image and text), cross-modal reconstruction (predicting one modality from another), masked modeling (learning contextual cross-modal embeddings), and instruction tuning (adapting multimodal reasoning to natural language instructions) \cite{yin2024survey}.
These strategies allow MFMs to generalize across perception tasks such as image captioning and speech-to-text translation, as well as reasoning tasks including visual question answering and multimodal dialogue. Their growing influence signals an important trend: foundation models are evolving from unimodal encoders into integrated multimodal systems that support increasingly rich forms of human–AI interaction.

\section{Metrics for Representation Alignment}
\label{metrics}

In this survey, we review the capacity of unimodal foundation models to achieve representation alignment, with a focus on how their learned representations behave across architectures and modalities. 
The central goal is to uncover potential commonalities and similarities among representations learned by different models and to evaluate the extent to which unimodal foundation models converge in their representation spaces.

Formally, let $\mathbf{X} = \{ \mathbf{x}_1, \dots, \mathbf{x}_n \} \in \mathbb{R}^{n \times d_1}$ and $\mathbf{Y} = \{ \mathbf{y}_1, \dots, \mathbf{y}_n \} \in \mathbb{R}^{n \times d_2}$ denote two sets of representations, extracted from distinct neural network layers or from different foundation models. 
Here, $n$ is the number of samples, and $d_1$ and $d_2$ are the feature dimensionalities. 
The central question is whether $\mathbf{X}$ and $\mathbf{Y}$ encode similar information, possibly up to admissible transformations such as rotation, scaling, or affine mapping.
To address this question, we first review representative similarity metrics that have been widely adopted in representation alignment analysis. 
These metrics provide principled tools for quantifying alignment quality. 

\subsection{Centered Kernel Alignment}
Centered kernel alignment (CKA) \cite{kornblith2019similarity, davari2023reliability} compares two representation sets by measuring the similarity of their kernel (Gram) matrices, which capture pairwise relationships between samples. 
We denote $\mathbf{K} = \mathbf{X}\mathbf{X}^\top$ and $\mathbf{L} = \mathbf{Y}\mathbf{Y}^\top$ as the linear Gram (kernel) matrices of $\mathbf{X}$ and $\mathbf{Y}$, which represent inner products between samples in the respective feature spaces. 
Typically, CKA first centers both kernel matrices to remove the influence of mean offsets:
\begin{equation}
\tilde{\mathbf{K}} = \mathbf{H}\mathbf{K}\mathbf{H}, \  \tilde{\mathbf{L}} = \mathbf{H}\mathbf{L}\mathbf{H}.
\end{equation}
where $\mathbf{H} = \mathbf{I}_n - \frac{1}{n}\mathbf{1}_n\mathbf{1}_n^\top$ denotes the centering matrix, $\mathbf{I}_n \in \mathbb{R}^{n\times n}$ is  the identity matrix, and $\mathbf{1}_n \in \mathbb{R}^{n}$ denotes the all-ones column vector.
Then, the linear CKA between $\mathbf{X}$ and $\mathbf{Y}$ can be defined as:
\begin{equation}
\mathrm{CKA}(\mathbf{X},\mathbf{Y})= \frac{\mathtt{HSIC}(\mathbf{K}, \mathbf{L})}{\sqrt{\mathtt{HSIC}(\mathbf{K}, \mathbf{K}) \mathtt{HSIC}(\mathbf{L}, \mathbf{L})}},
\end{equation}
where $\mathtt{HSIC}(\cdot, \cdot)$ denotes the Hilbert-Schmidt Independence Criterion (HSIC) that measures the dependence between the two kernel spaces:
\begin{equation}
\mathtt{HSIC}(\mathbf{K}, \mathbf{L}) = \mathrm{tr}(\tilde{\mathbf{K}} \tilde{\mathbf{L}}).
\end{equation}

CKA normalizes HSIC to produce a scale-invariant similarity measure. It can be interpreted as the cosine of the angle between the centered kernel matrices $\tilde{K}$ and $\tilde{L}$, when viewed as elements in the Hilbert-Schmidt space. 
It is invariant to isotropic scaling, i.e., $\mathrm{CKA}(c\mathbf{X},\mathbf{Y}) = \mathrm{CKA}(\mathbf{X},\mathbf{Y})$ for any scalar $c \neq 0$, and to orthogonal transformations, i.e., $\mathrm{CKA}(\mathbf{X}\mathbf{Q},\mathbf{Y}) = \mathrm{CKA}(\mathbf{X},\mathbf{Y})$ for any orthogonal matrix $\mathbf{Q}$.
The resulting CKA score ranges from $0$ to $1$, with $1$ indicating perfect alignment between $\mathbf{X}$ and $\mathbf{Y}$.
Based on CKA, several variants have been proposed to enhance its robustness or adapt it to specific settings, such as unbiased CKA \cite{song2007supervised}, kernel CKA \cite{kornblith2019similarity}, and class-conditional CKA \cite{nguyen2021do}.
These refinements have made CKA one of the most widely used metrics for comparing neural representations across architectures and training settings.

\subsection{Canonical Correlation Analysis}
Canonical correlation analysis (CCA) \cite{Hotelling, morcos2018insights} is a classical statistical technique that identifies linear relationships between two multivariate datasets.
It seeks linear projections of two random vectors such that their resulting projected representations are maximally correlated. 
Assume that both input matrices $\mathbf{X}$ and $\mathbf{Y}$ are centered (i.e., each column has zero mean). 
CCA aims to find directions $\mathbf{a} \in \mathbb{R}^{d_1}$ and $\mathbf{b} \in \mathbb{R}^{d_2}$ such that the projections $\mathbf{X}\mathbf{a}$ and $\mathbf{Y}\mathbf{b}$ are maximally correlated. This is formalized as:
\begin{equation}
\max_{\mathbf{a}, \mathbf{b}} \ \rho = \frac{\mathbf{a}^\top \mathbf{C}_{XY} \mathbf{b}}{
\sqrt{\mathbf{a}^\top \mathbf{C}_{XX} \mathbf{a}} \cdot \sqrt{\mathbf{b}^\top \mathbf{C}_{YY} \mathbf{b}}}
\end{equation}
where $\mathbf{C}_{XX} = \frac{1}{n} \mathbf{X}^\top \mathbf{X}$ and $\mathbf{C}_{YY} = \frac{1}{n} \mathbf{Y}^\top \mathbf{Y}$ 
are the covariance matrices of $\mathbf{X}$ and $\mathbf{Y}$, respectively, and $\mathbf{C}_{XY} = \frac{1}{n} \mathbf{X}^\top \mathbf{Y}$ is the corresponding  cross-covariance matrix.
The solution yields the first pair of canonical variates:
\begin{equation}
\mathbf{u}_1 = \mathbf{X}\mathbf{a}_1, \quad \mathbf{v}_1 = \mathbf{Y}\mathbf{b}_1,
\end{equation}
where $\rho_1 = \mathrm{corr}(\mathbf{u}_1, \mathbf{v}_1)$ is the largest (first) canonical correlation.
Subsequent pairs $(\mathbf{a}_i, \mathbf{b}_i)$ can be derived in a similar manner, subject to the orthogonality constraints:
\begin{equation}
\mathbf{u}_i^\top \mathbf{u}_j = 0, \quad \mathbf{v}_i^\top \mathbf{v}_j = 0 \quad \text{for} \quad  \forall i \ne j.
\end{equation}
The number of nonzero canonical correlations is at most $r = \min(d_1, d_2)$, and the sequence $\rho_1 \ge \rho_2 \ge \dots \ge \rho_r \ge 0$ quantifies the strength of the linear relationship between $\mathbf{X}$ and $\mathbf{Y}$.

An important extension is singular vector CCA (SVCCA) \cite{raghu2017svcca, artetxes}, which first reduces both $\mathbf{X}$ and $\mathbf{Y}$ to their dominant subspaces using singular value decomposition and then applies CCA.
This improves robustness to noise and has become a standard technique for comparing deep learning representations. Both CCA and SVCCA are invariant to affine transformations, while SVCCA further filters out low-variance directions, improving stability in practice.

\subsection{Mutual Nearest Neighbors}
Mutual nearest neighbors (MNN) \cite{haghverdi2018batch} define a symmetric relationship between samples from two sets and are commonly used to establish robust correspondences between learned representations. 
Let $\mathtt{NN}_k(\mathbf{x}_i; \mathbf{Y})$ denote the set of $k$ nearest neighbors of $\mathbf{x}_i \in \mathbf{X}$ in $\mathbf{Y}$, measured under a specific distance metric.
A pair $(\mathbf{x}_i, \mathbf{y}_j)$ is said to form a mutual nearest neighbor pair if:
\begin{equation}
\mathbf{y}_j \in \mathtt{NN}_k(\mathbf{x}_i; \mathbf{Y}) \quad \text{and} \quad \mathbf{x}_i \in \mathtt{NN}_k(\mathbf{y}_j; \mathbf{X}).
\end{equation}
MNN is widely used to reduce false-positive matches, particularly in high-dimensional or noisy representation spaces.
Several variants have been developed based on the MNN principle. For example, mutual k-nearest neighbor matching \cite{Platonic} has been applied to evaluate cross-modal  representation alignment.

The selection of metric depends on the specific aspect of similarity one seeks to capture. In practice, CKA is often preferred for its robustness and interpretability across architectures and tasks, whereas CCA provides useful insights when comparing closely related spaces, and MNN proves valuable when local semantic structures are of interest.
Beyond CKA, CCA, and MNN, a number of additional metrics have been developed to capture complementary aspects of representation similarity. Examples include Riemannian distance \cite{shahbazi2021using}, which accounts for the geometry of covariance matrices; similarity-of-similarity matrices (SSM) \cite{diedrichsen2017representational}, which measure agreement in pairwise similarity structures; and rank-based or Jaccard similarity metrics \cite{wang2020towards}, which focus on relational consistency. For a systematic overview of these approaches, refer to the paper by \citet{klabunde2025similarity}, which provides a detailed survey of similarity metrics for representation analysis.
\section{Representation Potentials of Foundation Models for Alignment}
\label{convergence}

In the following subsections, we review existing works in vision, language, speech, modalities, and neuroscience that explore the representation potentials of foundation models for alignment.

\subsection{Representation Alignment in Vision}
Within computer vision, a growing body of evidence suggests that models with different architectures, training objectives, and datasets can develop compatible understandings of visual information.
Early studies established that shallow features and early convolutional layers behave in similar ways. For example, \citet{lenc2015understanding} demonstrated that representations such as histograms of oriented gradients (HOG) and early convolutional filters respond linearly to geometric transformations like warps and flips, revealing that these early features are broadly interchangeable across architectures. \citet{li2015convergent} showed that independently trained networks often develop neuron clusters with overlapping functions, indicating partial convergence in learned representations.
\citet{Raghu20175} introduced SVCCA and reported that neural representations exhibit strong cross-initialization similarity, with lower layers converging early into compact shared subspaces while higher layers continue to evolve more gradually.
\citet{morcos2018insights} found that networks with better generalization exhibit higher representation similarity across random initializations, while overfitted networks diverge more.
\citet{kornblith2019similarity} provided systematic evidence that wider models learn more similar representations, early layers converge quickly, and deeper layers often contain redundancies across consecutive layers.

Subsequent works examined alignment under varied training objectives and architectures.
\citet{csiszarik2021similarity} showed that inner representations in deep convolutional networks with identical architectures but different initializations can be closely matched using only a single affine stitching layer.
\citet{roeder2021linear} proved that a broad class of discriminative and autoregressive models are identifiable in function space up to a linear transformation.
\citet{grigg2021self} compared supervised and self-supervised training, finding that intermediate layers are strikingly similar across paradigms, but final layers diverge: supervised models emphasize class-specific structure, whereas self-supervised models emphasize invariance to augmentations.
\citet{bansal2021revisiting} introduced the concept of stitching connectivity, showing that identically structured networks trained in different ways can be stitched together at intermediate layers with minimal performance degradation.
\citet{caron2021emerging} highlighted how self-supervised Vision Transformers (ViTs) consistently converge to similar spatial attention patterns and semantic structures, regardless of the specific training setup.
\citet{raghu2021vision} compared CNNs and ViTs using CKA, finding divergence in early layers but convergence in later ones. 
\citet{moschella2022relative} noted that although absolute coordinates of latent embeddings vary across training runs, relative angular relationships are preserved, reflecting alignment at a relational level.

More recent studies have reinforced and expanded these findings.
\citet{shekhar2023objectives} reported that models trained with the same self-supervised objective tend to learn more similar representations, even when the architectures differ.
\citet{dinov2} showed that self-supervised ViTs trained on different datasets or initializations learn similar high-level visual structures and that their features are compatible with those of supervised models.
\citet{dravid2023rosetta} identified Rosetta neurons, which reliably emerge across model architectures, training paradigms, and tasks.
\citet{stoica2024zipit} demonstrated that independently trained networks can be merged without retraining by aligning and zipping their feature spaces.
\citet{sharon2025} showed that during training, representation similarity exhibits distinct phases whose clarity depends on both architecture and optimizer: SGD and ViTs exhibit more synchronized, sharply delineated evolution of layer representations, whereas ResNets and Adam yield more gradual or less aligned dynamics.
\citet{li2025dual} highlighted that bidirectional Transformers serve as strong representation learners, enabling unified modeling of multimodal data distributions through likelihood estimation.

Beyond classification, alignment has also been observed in generative contexts.
\citet{yu2025representation} argued that the success of diffusion-based generation hinges on learning meaningful representations, and proposed a regularization strategy that aligns noisy denoising states with clean embeddings from pretrained encoders.
Moreover, several studies \cite{balestriero2018spline, kornblith2019similarity, roeder2021linear, Platonic} converge on a broader trend: representation similarity increases with model scale and performance. 
In other words, as vision models become larger, more expressive, and more generalizable, their internal representations tend to align more closely, pointing to a fundamental convergence.

\subsection{Representation Alignment in Language}
LLMs are increasingly demonstrating human-level proficiency across a broad spectrum of natural language processing tasks, including knowledge extraction, reasoning, and dialogue. This widespread improvement suggests a potential convergence in how these models process and represent linguistic information.
A growing body of research provides evidence that diverse LLMs, often trained with different architectures and datasets, nonetheless develop aligned internal representations.
One line of work highlights consistent structural patterns.
\citet{phang2021fine} found a block-structured similarity pattern in the hidden representations of fine-tuned RoBERTa \cite{roberta} and ALBERT \cite{Lan2020ALBERT}, suggesting that training induces stable, repeatable alignment across models.
\citet{jiang2025tracing} similarly observed that representation similarity in transformer models is strongest between adjacent layers, pointing to a layerwise convergence mechanism.

Another direction emphasizes concept-level alignment. 
\citet{park2024linear} formalized what it means for high-level concepts to be linearly represented in LLMs, introduced a causal inner product to capture semantic separability, and showed that high-level concepts in LLaMA2 \cite{touvron2023llama} can be probed or steered as approximate linear directions.
\citet{lan2024quantifying} decomposed LLM activations with sparse autoencoders, revealing disentangled features that align closely across different models.
\citet{burger2024truth} reported the emergence of a universal two-dimensional truth representation across LLMs of varying sizes and architectures, while \citet{tan2024analysing} identified strong correlations in both in-distribution and out-of-distribution steerability between LLaMA \cite{touvron2023llama} and Qwen \cite{qwen}.

A third body of evidence focuses on transferable features and universal neurons.
\citet{del2022cross} proposed a neuron-wise correlation metric that reveals the "first align, then predict" pattern across languages in multilingual models more faithfully.
\citet{gurnee2024universal} showed that about one to five percent of neurons in independently seeded GPT2 models are universal, interpretable, and causally relevant for predictions.
\citet{oozeer2025} found that safety-intervention vectors discovered in the activation space of one LLM can be mapped into the activation spaces of other LLMs via learned autoencoder mappings.
\citet{chen2025transferring} find that affine mappings between residual streams allow for effective transfer of learned feature modules, probes, and steering vectors from small to large models.
\citet{rinaldi2025update} further demonstrated that task vectors can be transferred from older to newer models without data or retraining by aligning weight structures of the two pretrained models.
\citet{lee2025shared} showed that token embeddings within a model family share both global and local geometry, enabling cross-model steering despite dimensional differences.

Finally, several studies point to broader universality across model classes.
\citet{wang2025towards} compared Transformer and Mamba models \cite{gu2024mamba} trained on the same data and found that they share many internal features and circuits, suggesting substantial but imperfect universality of mechanisms across architectures.
\citet{cheng2025emergence} analyzed intrinsic dimensionality of internal representations in transformer-based language models and identify a high-dimensional abstraction phase within the middle layers. 
This phase, consistently observed across architectures and datasets, reflects the point where models begin to form abstract, task-relevant representations that generalize well across tasks and models.

\subsection{Representation Alignment in Speech}
The study of speech models, particularly those trained using self-supervised learning (SSL) \cite{mohamed2022self, riera2023phone}, also reveals emerging evidence of representation alignment.
For instance, 
\citet{ollerenshaw21} examined end-to-end automatic speech recognition systems and found that CNN-based models exhibited progressively hierarchical and stable representation similarity as depth increased, whereas LSTM and Transformer architectures displayed less clean or more irregular similarity structures across layers. 
\citet{chung2021similarity} reported that the choice of learning objective has a larger impact on how similar representations are across models than architectural choices.
\citet{pasad2023co, pasad2024self} provided a detailed analysis of how acoustic, phonetic, and word-level information emerge at different layers, showing that both pretraining objectives and model size determine where key linguistic properties such as identity, pronunciation, syntax, and semantics are encoded.
\citet{waheed2024speech} showed that certain SFMs achieve strong zero-shot performance on tasks for which they were never explicitly trained, and that this performance correlates with higher-quality underlying representations.
\citet{dorszewski25} found that Transformer-based speech representation models exhibit a block-structured similarity pattern across layers, with substantial redundancy within blocks.
\citet{huo2025ite} demonstrated that the contrast between HuBERT \cite{hsu2021hubert} and wav2vec2.0 \cite{wav2vec2} lies not in whether a contrastive or classification objective is used, but in the iterative pseudo-label refinement strategy: multiple clustering iterations yield more aligned representations.

\subsection{Representation Alignment Across Modalities}
\label{cross}

Beyond alignment within individual modalities, an increasing number of studies highlight the potential for foundation model representations to align across different modalities, even when trained independently. 
One line of studies shows that representations from language and vision models can be brought into correspondence with lightweight mappings.
For example, 
\citet{linearly} investigated the extent to which conceptual representations from frozen text-only and vision-only models align, and found that visual information encoded by image models can be transferred to language models using only a single learned linear projection.
Similarly, \citet{LtoI} proposed a lightweight framework for adapting pretrained text-only LLMs to handle multimodal inputs. Their approach enables LLMs to process interleaved image–text data and generate text interleaved with retrieved images, all without retraining the base model.
\citet{Maniparambil} examined whether inherent alignment exists between independently trained unimodal vision and language encoders, and found that such models encode semantically similar structures, enabling zero-shot latent space communication without explicit alignment.
\citet{zhang2024assessing} evaluated the degree to which independently trained vision and language models can be aligned and propose an efficient alignment framework for downstream tasks.

Other studies explore alignment between language and auditory representations.
\citet{ngo2024language} showed that auditory and language representations can be approximately aligned through a simple linear transformation, pointing to a shared structural basis across modalities.
They further demonstrated that text-only LLMs encode features that align with auditory object representations, such that a contrastive probe can successfully retrieve the correct object label from an audio snippet.
\citet{lee2024multimodal} revealed that cross-modal representations, particularly between text and speech, tend to converge in the deeper layers of the models, while the early layers remain modality-specific and specialized for raw input processing. This layer-wise convergence reflects a progressive transformation toward modality-agnostic abstraction as signals propagate through the network.

These results resonate with the Platonic Representation Hypothesis \cite{Platonic}, which argues that foundation models are converging toward a common statistical model of reality embedded within their representation spaces.
All these findings suggest that cross-modal convergence may reflect a deeper tendency of foundation models to discover modality-agnostic abstractions. This emerging evidence points to the possibility that foundation models, even when trained separately, may inhabit overlapping representational manifolds that facilitate alignment, transfer, and integration across modalities.

\subsection{Representation Alignment with Neuroscience}
While the previous sections focus on alignment within and across artificial modalities, a natural question is whether the representations learned by foundation models also correspond to those observed in biological systems. 
In this context, representation alignment does not refer to internal consistency within neuroscience, but rather to the similarity and correspondence between model-derived features and neural representations measured in cognitive neuroscience \cite{biology12101330}. This perspective bridges artificial and biological intelligence, providing insights into the extent to which foundation models capture structures present in natural cognition.
For example,
\citet{chen2024se} showed that both Wav2Vec2.0 and GPT2 predict human auditory cortex responses, with model activations exhibiting strong correlations to brain activity during speech and language perception.
\citet{khosla2024privileged} developed axis-sensitive metrics for alignment and demonstrated that both biological and artificial neural networks exhibit privileged, non-arbitrary axes of representation that converge across systems.
\citet{hosseini2024universality} examined the representation universality hypothesis, and proposed that artificial neural networks trained on naturalistic data converge toward shared representational structures that also align with the brain.
Their findings show that inter-model representational agreement reliably predicts brain alignment, implying that shared tasks and environments naturally drive both artificial and biological systems toward similar representational structures.
\citet{DoerigHigh} reported that embeddings from LLMs of whole scene captions align closely with high-level visual cortex responses to corresponding images, outperform many image-based models, and that image models trained to predict these caption embeddings can match or exceed vision model alignment to brain activity.
\citet{raugel2025disentangling} added that model size, training scale, and the nature of image data all critically drive how well vision transformer models develop representations aligned with human brain activity, with larger, more human-centric models aligning later brain regions and temporal dynamics when given enough training.
\citet{feather2025brain} proposed the NeuroAI Turing Test, a benchmark that requires models to match both behavior and internal neural representations of brains, arguing that this stronger criterion is necessary for model-brain evaluations in NeuroAI.
Taken together, these findings suggest that artificial and biological systems, despite differences in architecture and learning mechanisms, converge toward similar representational frameworks when exposed to comparable sensory inputs and functional objectives. This bridges the gap between artificial intelligence and neuroscience, offering insights into both machine learning and human cognition.

\section{Factors Driving Representation Potential for  Alignment}
\label{factors}
A variety of factors contribute to the alignment potential of foundation model representations. Among the most prominent is scale, which encompasses model capacity, dataset size, and computational resources.
\citet{kaplan2020scaling} established scaling laws showing that performance improves predictably with these factors, and subsequent work has demonstrated their impact on representation similarity.
For instance, \citet{gammelgaard2023large} showed that as LLMs increase in size and quality, they organize concepts in embedding spaces in ways increasingly similar to the structures of knowledge graphs, suggesting convergence toward human-like conceptual organization derived from text alone.
\citet{nguyen2024multilingual} demonstrated that linguistic and cultural diversity in data enhances generalization and robustness.
\citet{Platonic} proposed that across multiple modalities, larger models trained on more diverse datasets naturally develop more aligned representations.

While the Platonic Representation
Hypothesis \cite{Platonic} suggests that convergence may occur independently of architectural or training objectives, evidence shows that these factors can nonetheless shape the dynamics of alignment.
The Transformer architecture \cite{NIPS2017_3f5ee243}, now dominant in foundation models, is argued to inherently facilitate generalizable representations due to its flexible inductive biases \cite{edelman2022inductive, bhattamis, geerts2025relational}. 
Training paradigms are equally influential.
Self-supervision, in particular, has been shown to encourage representations that generalize broadly across tasks and domains.
\citet{ciernik2024training} found that self-supervised vision models yield stronger pairwise similarity generalization across datasets compared to models trained with image classification or image–text matching objectives. 
These results indicate that although convergence may not depend strictly on architecture or objective, both exert important shaping effects on representational outcomes.

The generality of tasks and instructions also affect representation potentials. 
As foundation models are trained on increasingly diverse task mixtures and fine-tuned with a wide range of instructions, their representation spaces become progressively constrained, encouraging the development of task-agnostic abstractions that potentially support alignment. 
\citet{sanh2022multitask} found that prompt-based multitask fine-tuning yields strong zero-shot generalization, frequently exceeding the performance of larger models.
\citet{chung2024scaling} further showed that scaling instruction-finetuning by expanding both the number and variety of finetuning tasks substantially boosts performance across zero-shot, few-shot, reasoning, multilingual, and open-ended benchmarks.
\citet{zhang24instruction} highlighted that instruction diversity, rather than the sheer number of examples per instruction, is the critical factor for driving generalization.

\section{Open Questions}
\label{sec7}

Despite compelling evidence for representation potentials of foundation models for alignment, several limitations and open questions remain.
One fundamental limitation arises from differences across modalities. 
Distinct sensors and perspectives capture complementary aspects of reality, and certain information may be unique to a given sensory channel, which constrains the extent to which their representations can perfectly align.
For example, visual data emphasizes spatial and perceptual detail, while language conveys abstract concepts and relationships. 
Consequently, full convergence to a single, identical representation across modalities is neither achievable nor necessarily desirable.
In this sense, alignment should perhaps be understood not as perfect overlap but as the development of partially shared abstractions that remain complementary across modalities.
Notably, \citet{lu2025indra} recently proposed that foundation models converge toward relational, externally grounded representations, implicitly reflecting a shared relational structure underlying reality. 
Such representations inherently involve contextualization and mutual reference across samples, and thus offer a principled way of addressing the limitations outlined above.

Another challenge lies in how to evaluate alignment rigorously.
While various metrics have been proposed to quantify the similarity between representations, an ongoing debate persists within the research community regarding the effectiveness and interpretability of these measures.
It is often unclear whether a given alignment score indicates a strong degree of similarity with only minor discrepancies, or a relatively weak alignment with significant underlying differences that are yet to be fully understood. 
For example, \citet{wang2018towards} introduced a neuron activation subspace match model to define the similarity between networks trained with the same architecture but different random initializations. They showed that convolutional layers often exhibit low similarity across independently trained networks, even with identical architectures.
\citet{brown2023} demonstrated that measures such as model stitching and CKA can reveal internal differences in language models that are invisible to performance-based evaluations.
\citet{harvey24a} provided a decoder-based perspective, arguing that high similarity in metrics like CKA or CCA implies that many features can be decoded similarly, but the reverse does not necessarily hold: models may allow similar decoding for some tasks while still differing geometrically in how information is distributed across dimensions. 
The absence of a universally accepted standard complicates cross-study comparisons and raises questions about what alignment scores truly capture.

The role of data bias and sociotechnical context also presents a crucial consideration. 
Training data are often scraped from the internet and thus inherit the biases, cultural norms, and imbalances embedded in human human-generated content.
Such biases shape representation spaces and constrain the universality of alignment claims. Beyond data, the broader sociotechnical context, including who builds the models and for what purposes, also influences alignment, raising questions about whether observed consistencies reflects general cognitive structures or artifacts of specific training regimes.

Finally, there are counterexamples and scenarios where representation alignment might not emerge. Highly specialized models, optimized for narrow tasks, may develop unique representations that diverge from general-purpose abstractions. In domains such as robotics, where standardized representations for complex sensorimotor experiences are still under development, 
the potential for alignment may be constrained by the absence of common frameworks and data formats.
In summary, while the evidence for alignment across architectures and modalities is strong, the boundaries of this phenomenon remain poorly understood. Future progress requires both more robust evaluation frameworks and a careful recognition of the contexts and biases that shape representational spaces.

\section{Conclusion}
\label{conclusion}
In this survey, we synthesize substantial evidence for the representation potential of foundation models, demonstrating their capacity for alignment within individual modalities such as vision, language, and speech, across multimodal combinations, and in relation to representations observed in neuroscience.
We analyze the key factors that
foster representation potentials and discuss open questions.
Our future work will pursue a deeper theoretical grounding of representation potentials.
We believe that continued exploration in this direction will not only drive the development of more interpretable foundation models but also enrich our broader understanding of the principles that underlie both artificial and natural intelligence.
\section*{Limitations}

This survey has focused on the representation potentials of foundation models across vision, language, speech, multimodality, and neuroscience, emphasizing areas where substantial empirical evidence is available.
However, our analysis necessarily excludes domains such as robotics, sensorimotor control, and graphs, where research on representation alignment is still fragmented and publicly available findings are limited.
As a result, this survey may not fully reflect the breadth of representational behaviors across all applications of foundation models.

Another limitation arises from the evaluation of alignment itself. While we reviewed a range of commonly used metrics, including CKA, CCA, and MNN, the field still lacks a unified standard for assessing representation similarity. This makes it difficult to integrate results across studies rigorously. Comparisons across modalities, architectures, or training objectives therefore remain partly qualitative, and definitive meta-analyses are constrained by methodological inconsistencies.

Finally, foundation models are rapidly evolving. New models, training paradigms, and evaluation techniques continue to emerge, particularly in cross-modal and neuroscience-inspired settings. Consequently, the conclusions and scope presented here should be regarded as a reflection of the current state of the field rather than a definitive account. We anticipate that many of the questions identified in this survey will be revisited and refined as the field advances.

\bibliography{custom}

\begin{thebibliography}{139}
\providecommand{\natexlab}[1]{#1}

\bibitem[{Achiam et~al.(2023)Achiam, Adler, Agarwal, Ahmad, Akkaya, Aleman, Almeida, Altenschmidt, Altman, Anadkat et~al.}]{gpt4}
Josh Achiam, Steven Adler, Sandhini Agarwal, Lama Ahmad, Ilge Akkaya, Florencia~Leoni Aleman, Diogo Almeida, Janko Altenschmidt, Sam Altman, Shyamal Anadkat, and 1 others. 2023.
\newblock Gpt-4 technical report.
\newblock \emph{arXiv preprint arXiv:2303.08774}.

\bibitem[{Alayrac et~al.(2022)Alayrac, Donahue, Luc, Miech, Barr, Hasson, Lenc, Mensch, Millican, Reynolds et~al.}]{Flamingo}
Jean-Baptiste Alayrac, Jeff Donahue, Pauline Luc, Antoine Miech, Iain Barr, Yana Hasson, Karel Lenc, Arthur Mensch, Katherine Millican, Malcolm Reynolds, and 1 others. 2022.
\newblock Flamingo: a visual language model for few-shot learning.
\newblock \emph{Advances in neural information processing systems}, 35:23716--23736.

\bibitem[{Artetxe et~al.(2020)Artetxe, Ruder, and Yogatama}]{artetxes}
Mikel Artetxe, Sebastian Ruder, and Dani Yogatama. 2020.
\newblock On the cross-lingual transferability of monolingual representations.
\newblock In \emph{Proceedings of the 58th Annual Meeting of the Association for Computational Linguistics}, pages 4623--4637.

\bibitem[{Azad et~al.(2023)Azad, Azad, Eskandari, Bozorgpour, Kazerouni, Rekik, and Merhof}]{azad2023foundational}
Bobby Azad, Reza Azad, Sania Eskandari, Afshin Bozorgpour, Amirhossein Kazerouni, Islem Rekik, and Dorit Merhof. 2023.
\newblock Foundational models in medical imaging: A comprehensive survey and future vision.
\newblock \emph{arXiv preprint arXiv:2310.18689}.

\bibitem[{Baevski et~al.(2020)Baevski, Zhou, Mohamed, and Auli}]{wav2vec2}
Alexei Baevski, Yuhao Zhou, Abdelrahman Mohamed, and Michael Auli. 2020.
\newblock wav2vec 2.0: A framework for self-supervised learning of speech representations.
\newblock \emph{Advances in neural information processing systems}, 33:12449--12460.

\bibitem[{Bai et~al.(2023)Bai, Bai, Chu, Cui, Dang, Deng, Fan, Ge, Han, Huang et~al.}]{qwen}
Jinze Bai, Shuai Bai, Yunfei Chu, Zeyu Cui, Kai Dang, Xiaodong Deng, Yang Fan, Wenbin Ge, Yu~Han, Fei Huang, and 1 others. 2023.
\newblock Qwen technical report.
\newblock \emph{arXiv preprint arXiv:2309.16609}.

\bibitem[{Balestriero and richard baraniuk(2018)}]{balestriero2018spline}
Randall Balestriero and richard baraniuk. 2018.
\newblock A spline theory of deep learning.
\newblock In \emph{Proceedings of the 35th International Conference on Machine Learning}, volume~80 of \emph{Proceedings of Machine Learning Research}, pages 374--383. PMLR.

\bibitem[{Bansal et~al.(2021)Bansal, Nakkiran, and Barak}]{bansal2021revisiting}
Yamini Bansal, Preetum Nakkiran, and Boaz Barak. 2021.
\newblock Revisiting model stitching to compare neural representations.
\newblock \emph{Advances in neural information processing systems}, 34:225--236.

\bibitem[{Barrault et~al.(2023)Barrault, Chung, Meglioli, Dale, Dong, Duquenne, Elsahar, Gong, Heffernan, Hoffman et~al.}]{barrault2023seamlessm4t}
Lo{\"\i}c Barrault, Yu-An Chung, Mariano~Cora Meglioli, David Dale, Ning Dong, Paul-Ambroise Duquenne, Hady Elsahar, Hongyu Gong, Kevin Heffernan, John Hoffman, and 1 others. 2023.
\newblock Seamlessm4t: Massively multilingual \& multimodal machine translation.
\newblock \emph{arXiv preprint arXiv:2308.11596}.

\bibitem[{Belinkov and Glass(2017)}]{belinkov2017}
Yonatan Belinkov and James Glass. 2017.
\newblock Analyzing hidden representations in end-to-end automatic speech recognition systems.
\newblock \emph{Advances in Neural Information Processing Systems}, 30.

\bibitem[{Bhattamishra et~al.(2023)Bhattamishra, Patel, Kanade, and Blunsom}]{bhattamis}
Satwik Bhattamishra, Arkil Patel, Varun Kanade, and Phil Blunsom. 2023.
\newblock Simplicity bias in transformers and their ability to learn sparse {B}oolean functions.
\newblock In \emph{Proceedings of the 61st Annual Meeting of the Association for Computational Linguistics (Volume 1: Long Papers)}, pages 5767--5791. Association for Computational Linguistics.

\bibitem[{Bommasani et~al.(2021)Bommasani, Hudson, Adeli, Altman, Arora, von Arx, Bernstein, Bohg, Bosselut, Brunskill et~al.}]{bommasani2021opportunities}
Rishi Bommasani, Drew~A Hudson, Ehsan Adeli, Russ Altman, Simran Arora, Sydney von Arx, Michael~S Bernstein, Jeannette Bohg, Antoine Bosselut, Emma Brunskill, and 1 others. 2021.
\newblock On the opportunities and risks of foundation models.
\newblock \emph{arXiv preprint arXiv:2108.07258}.

\bibitem[{Brohan et~al.(2022)Brohan, Brown, Carbajal, Chebotar, Dabis, Finn, Gopalakrishnan, Hausman, Herzog, Hsu et~al.}]{brohan2022rt}
Anthony Brohan, Noah Brown, Justice Carbajal, Yevgen Chebotar, Joseph Dabis, Chelsea Finn, Keerthana Gopalakrishnan, Karol Hausman, Alex Herzog, Jasmine Hsu, and 1 others. 2022.
\newblock Rt-1: Robotics transformer for real-world control at scale.
\newblock \emph{arXiv preprint arXiv:2212.06817}.

\bibitem[{Brown et~al.(2023)Brown, Godfrey, Konz, Tu, and Kvinge}]{brown2023}
Davis Brown, Charles Godfrey, Nicholas Konz, Jonathan Tu, and Henry Kvinge. 2023.
\newblock Understanding the inner-workings of language models through representation dissimilarity.
\newblock In \emph{Proceedings of the 2023 Conference on Empirical Methods in Natural Language Processing}, pages 6543--6558. Association for Computational Linguistics.

\bibitem[{Brown et~al.(2020)Brown, Mann, Ryder, Subbiah, Kaplan, Dhariwal, Neelakantan, Shyam, Sastry, Askell, Agarwal, Herbert-Voss, Krueger, Henighan, Child, Ramesh, Ziegler, Wu, Winter, Hesse, Chen, Sigler, Litwin, Gray, Chess, Clark, Berner, McCandlish, Radford, Sutskever, and Amodei}]{GPT3}
Tom Brown, Benjamin Mann, Nick Ryder, Melanie Subbiah, Jared~D Kaplan, Prafulla Dhariwal, Arvind Neelakantan, Pranav Shyam, Girish Sastry, Amanda Askell, Sandhini Agarwal, Ariel Herbert-Voss, Gretchen Krueger, Tom Henighan, Rewon Child, Aditya Ramesh, Daniel Ziegler, Jeffrey Wu, Clemens Winter, and 12 others. 2020.
\newblock Language models are few-shot learners.
\newblock In \emph{Advances in Neural Information Processing Systems}, pages 1877--1901.

\bibitem[{B{\"u}rger et~al.(2024)B{\"u}rger, Hamprecht, and Nadler}]{burger2024truth}
Lennart B{\"u}rger, Fred~A Hamprecht, and Boaz Nadler. 2024.
\newblock Truth is universal: Robust detection of lies in llms.
\newblock \emph{Advances in Neural Information Processing Systems}, 37.

\bibitem[{Caron et~al.(2021)Caron, Touvron, Misra, J{\'e}gou, Mairal, Bojanowski, and Joulin}]{caron2021emerging}
Mathilde Caron, Hugo Touvron, Ishan Misra, Herv{\'e} J{\'e}gou, Julien Mairal, Piotr Bojanowski, and Armand Joulin. 2021.
\newblock Emerging properties in self-supervised vision transformers.
\newblock In \emph{Proceedings of the IEEE/CVF international conference on computer vision}, pages 9650--9660.

\bibitem[{Chen et~al.(2025)Chen, Merullo, Stolfo, and Pavlick}]{chen2025transferring}
Alan Chen, Jack Merullo, Alessandro Stolfo, and Ellie Pavlick. 2025.
\newblock Transferring features across language models with model stitching.
\newblock \emph{arXiv preprint arXiv:2506.06609}.

\bibitem[{Chen et~al.(2024)Chen, He, Fu, Fan, Chang, and Li}]{chen2024se}
Peili Chen, Linyang He, Li~Fu, Lu~Fan, Edward~F Chang, and Yuanning Li. 2024.
\newblock Do self-supervised speech and language models extract similar representations as human brain?
\newblock In \emph{ICASSP 2024-2024 IEEE International Conference on Acoustics, Speech and Signal Processing (ICASSP)}, pages 2225--2229. IEEE.

\bibitem[{Chen et~al.(2022{\natexlab{a}})Chen, Wang, Chen, Wu, Liu, Chen, Li, Kanda, Yoshioka, Xiao et~al.}]{chen2022wavlm}
Sanyuan Chen, Chengyi Wang, Zhengyang Chen, Yu~Wu, Shujie Liu, Zhuo Chen, Jinyu Li, Naoyuki Kanda, Takuya Yoshioka, Xiong Xiao, and 1 others. 2022{\natexlab{a}}.
\newblock Wavlm: Large-scale self-supervised pre-training for full stack speech processing.
\newblock \emph{IEEE Journal of Selected Topics in Signal Processing}, 16(6):1505--1518.

\bibitem[{Chen et~al.(2022{\natexlab{b}})Chen, Wang, Changpinyo, Piergiovanni, Padlewski, Salz, Goodman, Grycner, Mustafa, Beyer et~al.}]{chen2022pali}
Xi~Chen, Xiao Wang, Soravit Changpinyo, AJ~Piergiovanni, Piotr Padlewski, Daniel Salz, Sebastian Goodman, Adam Grycner, Basil Mustafa, Lucas Beyer, and 1 others. 2022{\natexlab{b}}.
\newblock Pali: A jointly-scaled multilingual language-image model.
\newblock \emph{arXiv preprint arXiv:2209.06794}.

\bibitem[{Cheng et~al.(2025)Cheng, Doimo, Kervadec, Macocco, Yu, Laio, and Baroni}]{cheng2025emergence}
Emily Cheng, Diego Doimo, Corentin Kervadec, Iuri Macocco, Lei Yu, Alessandro Laio, and Marco Baroni. 2025.
\newblock Emergence of a high-dimensional abstraction phase in language transformers.
\newblock In \emph{The Thirteenth International Conference on Learning Representations}.

\bibitem[{Chung et~al.(2024)Chung, Hou, Longpre, Zoph, Tay, Fedus, Li, Wang, Dehghani, Brahma et~al.}]{chung2024scaling}
Hyung~Won Chung, Le~Hou, Shayne Longpre, Barret Zoph, Yi~Tay, William Fedus, Yunxuan Li, Xuezhi Wang, Mostafa Dehghani, Siddhartha Brahma, and 1 others. 2024.
\newblock Scaling instruction-finetuned language models.
\newblock \emph{Journal of Machine Learning Research}, 25(70):1--53.

\bibitem[{Chung et~al.(2021)Chung, Belinkov, and Glass}]{chung2021similarity}
Yu-An Chung, Yonatan Belinkov, and James Glass. 2021.
\newblock Similarity analysis of self-supervised speech representations.
\newblock In \emph{ICASSP 2021-2021 IEEE International Conference on Acoustics, Speech and Signal Processing (ICASSP)}, pages 3040--3044. IEEE.

\bibitem[{Ciernik et~al.(2025)Ciernik, Linhardt, Morik, Dippel, Kornblith, and Muttenthaler}]{ciernik2024training}
Laure Ciernik, Lorenz Linhardt, Marco Morik, Jonas Dippel, Simon Kornblith, and Lukas Muttenthaler. 2025.
\newblock Objective drives the consistency of representational similarity across datasets.
\newblock In \emph{Forty-second International Conference on Machine Learning}.

\bibitem[{Csisz{\'a}rik et~al.(2021)Csisz{\'a}rik, K{\H{o}}r{\"o}si-Szab{\'o}, Matszangosz, Papp, and Varga}]{csiszarik2021similarity}
Adri{\'a}n Csisz{\'a}rik, P{\'e}ter K{\H{o}}r{\"o}si-Szab{\'o}, Akos Matszangosz, Gergely Papp, and D{\'a}niel Varga. 2021.
\newblock Similarity and matching of neural network representations.
\newblock \emph{Advances in Neural Information Processing Systems}, 34:5656--5668.

\bibitem[{Cui et~al.(2024)Cui, Yu, Jiao, Meng, Zhang, Wang, Guo, and King}]{cui2024recent}
Wenqian Cui, Dianzhi Yu, Xiaoqi Jiao, Ziqiao Meng, Guangyan Zhang, Qichao Wang, Yiwen Guo, and Irwin King. 2024.
\newblock Recent advances in speech language models: A survey.
\newblock \emph{arXiv preprint arXiv:2410.03751}.

\bibitem[{Cui et~al.(2022)Cui, Niekum, Gupta, Kumar, and Rajeswaran}]{cui2022can}
Yuchen Cui, Scott Niekum, Abhinav Gupta, Vikash Kumar, and Aravind Rajeswaran. 2022.
\newblock Can foundation models perform zero-shot task specification for robot manipulation?
\newblock In \emph{Learning for dynamics and control conference}, pages 893--905. PMLR.

\bibitem[{Davari et~al.(2023)Davari, Horoi, Natik, Lajoie, Wolf, and Belilovsky}]{davari2023reliability}
MohammadReza Davari, Stefan Horoi, Amine Natik, Guillaume Lajoie, Guy Wolf, and Eugene Belilovsky. 2023.
\newblock Reliability of {CKA} as a similarity measure in deep learning.
\newblock In \emph{The Eleventh International Conference on Learning Representations}.

\bibitem[{Del and Fishel(2022)}]{del2022cross}
Maksym Del and Mark Fishel. 2022.
\newblock Cross-lingual similarity of multilingual representations revisited.
\newblock \emph{arXiv preprint arXiv:2212.01924}.

\bibitem[{Devlin et~al.(2019)Devlin, Chang, Lee, and Toutanova}]{bert}
Jacob Devlin, Ming-Wei Chang, Kenton Lee, and Kristina Toutanova. 2019.
\newblock Bert: Pre-training of deep bidirectional transformers for language understanding.
\newblock In \emph{Proceedings of the 2019 conference of the North American chapter of the association for computational linguistics: human language technologies, volume 1}, pages 4171--4186.

\bibitem[{Diedrichsen and Kriegeskorte(2017)}]{diedrichsen2017representational}
J{\"o}rn Diedrichsen and Nikolaus Kriegeskorte. 2017.
\newblock Representational models: A common framework for understanding encoding, pattern-component, and representational-similarity analysis.
\newblock \emph{PLoS computational biology}, 13(4):e1005508.

\bibitem[{Doerig et~al.(2025)Doerig, Kietzmann, Allen, Wu, Naselaris, Kay, and Charest}]{DoerigHigh}
Adrien Doerig, \{Tim C.\} Kietzmann, Emily Allen, Yihan Wu, Thomas Naselaris, Kendrick Kay, and Ian Charest. 2025.
\newblock High-level visual representations in the human brain are aligned with large language models.
\newblock \emph{Nature Machine Intelligence}, 7(8):1220--1234.

\bibitem[{Dorszewski et~al.(2025)Dorszewski, Jacobsen, T{\v{e}}tkov{\'a}, and Hansen}]{dorszewski25}
Teresa Dorszewski, Albert~Kj{\o}ller Jacobsen, Lenka T{\v{e}}tkov{\'a}, and Lars~Kai Hansen. 2025.
\newblock How redundant is the transformer stack in speech representation models?
\newblock In \emph{ICASSP 2025-2025 IEEE International Conference on Acoustics, Speech and Signal Processing (ICASSP)}, pages 1--5. IEEE.

\bibitem[{Dosovitskiy et~al.(2021)Dosovitskiy, Beyer, Kolesnikov, Weissenborn, Zhai, Unterthiner, Dehghani, Minderer, Heigold, Gelly, Uszkoreit, and Houlsby}]{ViTs}
Alexey Dosovitskiy, Lucas Beyer, Alexander Kolesnikov, Dirk Weissenborn, Xiaohua Zhai, Thomas Unterthiner, Mostafa Dehghani, Matthias Minderer, Georg Heigold, Sylvain Gelly, Jakob Uszkoreit, and Neil Houlsby. 2021.
\newblock An image is worth 16x16 words: Transformers for image recognition at scale.
\newblock In \emph{International Conference on Learning Representations}.

\bibitem[{Dravid et~al.(2023)Dravid, Gandelsman, Efros, and Shocher}]{dravid2023rosetta}
Amil Dravid, Yossi Gandelsman, Alexei~A Efros, and Assaf Shocher. 2023.
\newblock Rosetta neurons: Mining the common units in a model zoo.
\newblock In \emph{Proceedings of the IEEE/CVF International Conference on Computer Vision}, pages 1934--1943.

\bibitem[{Edelman et~al.(2022)Edelman, Goel, Kakade, and Zhang}]{edelman2022inductive}
Benjamin~L Edelman, Surbhi Goel, Sham Kakade, and Cyril Zhang. 2022.
\newblock Inductive biases and variable creation in self-attention mechanisms.
\newblock In \emph{International Conference on Machine Learning}, pages 5793--5831. PMLR.

\bibitem[{Feather et~al.(2025)Feather, Khosla, Murty, and Nayebi}]{feather2025brain}
Jenelle Feather, Meenakshi Khosla, N~Murty, and Aran Nayebi. 2025.
\newblock Brain-model evaluations need the neuroai turing test.
\newblock \emph{arXiv preprint arXiv:2502.16238}.

\bibitem[{Firoozi et~al.(2023)Firoozi, Tucker, Tian, Majumdar, Sun, Liu, Zhu, Song, Kapoor, Hausman et~al.}]{firoozi2023foundation}
Roya Firoozi, Johnathan Tucker, Stephen Tian, Anirudha Majumdar, Jiankai Sun, Weiyu Liu, Yuke Zhu, Shuran Song, Ashish Kapoor, Karol Hausman, and 1 others. 2023.
\newblock Foundation models in robotics: Applications, challenges, and the future.
\newblock \emph{The International Journal of Robotics Research}.

\bibitem[{Gammelgaard et~al.(2023)Gammelgaard, Christiansen, and S{\o}gaard}]{gammelgaard2023large}
Mathias~Lykke Gammelgaard, Jonathan~Gabel Christiansen, and Anders S{\o}gaard. 2023.
\newblock Large language models converge toward human-like concept organization.
\newblock \emph{arXiv preprint arXiv:2308.15047}.

\bibitem[{Geerts et~al.(2025)Geerts, Chan, Clopath, and Stachenfeld}]{geerts2025relational}
Jesse Geerts, Stephanie Chan, Claudia Clopath, and Kimberly Stachenfeld. 2025.
\newblock Relational reasoning and inductive bias in transformers trained on a transitive inference task.
\newblock \emph{arXiv preprint arXiv:2506.04289}.

\bibitem[{Grattafiori et~al.(2024)Grattafiori, Dubey, Jauhri, Pandey, Kadian, and et~al.}]{llama3}
Aaron Grattafiori, Abhimanyu Dubey, Abhinav Jauhri, Abhinav Pandey, Abhishek Kadian, and Ahmad Al-Dahle et~al. 2024.
\newblock The llama 3 herd of models.

\bibitem[{Grigg et~al.(2021)Grigg, Busbridge, Ramapuram, and Webb}]{grigg2021self}
Tom~George Grigg, Dan Busbridge, Jason Ramapuram, and Russ Webb. 2021.
\newblock Do self-supervised and supervised methods learn similar visual representations?
\newblock \emph{arXiv preprint arXiv:2110.00528}.

\bibitem[{Gu and Dao(2024)}]{gu2024mamba}
Albert Gu and Tri Dao. 2024.
\newblock Mamba: Linear-time sequence modeling with selective state spaces.
\newblock In \emph{First Conference on Language Modeling}.

\bibitem[{Gurnee et~al.(2024)Gurnee, Horsley, Guo, Kheirkhah, Sun, Hathaway, Nanda, and Bertsimas}]{gurnee2024universal}
Wes Gurnee, Theo Horsley, Zifan~Carl Guo, Tara~Rezaei Kheirkhah, Qinyi Sun, Will Hathaway, Neel Nanda, and Dimitris Bertsimas. 2024.
\newblock Universal neurons in gpt2 language models.
\newblock \emph{arXiv preprint arXiv:2401.12181}.

\bibitem[{Ha and Schmidhuber(2018)}]{ha2018world}
David Ha and J{\"u}rgen Schmidhuber. 2018.
\newblock World models.
\newblock \emph{arXiv preprint arXiv:1803.10122}, 2(3).

\bibitem[{Haghverdi et~al.(2018)Haghverdi, Lun, Morgan, and Marioni}]{haghverdi2018batch}
Laleh Haghverdi, Aaron~TL Lun, Michael~D Morgan, and John~C Marioni. 2018.
\newblock Batch effects in single-cell rna-sequencing data are corrected by matching mutual nearest neighbors.
\newblock \emph{Nature biotechnology}, 36(5):421--427.

\bibitem[{Harvey et~al.(2024)Harvey, Lipshutz, and Williams}]{harvey24a}
Sarah~E Harvey, David Lipshutz, and Alex~H Williams. 2024.
\newblock What representational similarity measures imply about decodable information.
\newblock In \emph{Proceedings of UniReps: the Second Edition of the Workshop on Unifying Representations in Neural Models}, volume 285 of \emph{Proceedings of Machine Learning Research}, pages 140--151. PMLR.

\bibitem[{He et~al.(2016)He, Zhang, Ren, and Sun}]{he2016deep}
Kaiming He, Xiangyu Zhang, Shaoqing Ren, and Jian Sun. 2016.
\newblock Deep residual learning for image recognition.
\newblock In \emph{CVPR}, pages 770--778.

\bibitem[{He et~al.(2020)He, Liu, Gao, and Chen}]{deberta}
Pengcheng He, Xiaodong Liu, Jianfeng Gao, and Weizhu Chen. 2020.
\newblock Deberta: Decoding-enhanced bert with disentangled attention.
\newblock \emph{arXiv preprint arXiv:2006.03654}.

\bibitem[{Hosseini et~al.(2024)Hosseini, Casto, Zaslavsky, Conwell, Richardson, and Fedorenko}]{hosseini2024universality}
Eghbal Hosseini, Colton Casto, Noga Zaslavsky, Colin Conwell, Mark Richardson, and Evelina Fedorenko. 2024.
\newblock Universality of representation in biological and artificial neural networks.
\newblock \emph{bioRxiv}.

\bibitem[{Hotelling(1936)}]{Hotelling}
Harold Hotelling. 1936.
\newblock Relations between two sets of variates.
\newblock \emph{Biometrika}, 28(3/4):321--377.

\bibitem[{Hsu et~al.(2021)Hsu, Bolte, Tsai, Lakhotia, Salakhutdinov, and Mohamed}]{hsu2021hubert}
Wei-Ning Hsu, Benjamin Bolte, Yao-Hung~Hubert Tsai, Kushal Lakhotia, Ruslan Salakhutdinov, and Abdelrahman Mohamed. 2021.
\newblock Hubert: Self-supervised speech representation learning by masked prediction of hidden units.
\newblock \emph{IEEE/ACM transactions on audio, speech, and language processing}, 29:3451--3460.

\bibitem[{Huang et~al.(2024)Huang, Li, Zhou, Yang, Liu, Liang, Zheng, Zhang, and Wang}]{huang2024enhancing}
Weijian Huang, Cheng Li, Hong-Yu Zhou, Hao Yang, Jiarun Liu, Yong Liang, Hairong Zheng, Shaoting Zhang, and Shanshan Wang. 2024.
\newblock Enhancing representation in radiography-reports foundation model: A granular alignment algorithm using masked contrastive learning.
\newblock \emph{Nature Communications}, 15(1):7620.

\bibitem[{Huh et~al.(2024)Huh, Cheung, Wang, and Isola}]{Platonic}
Minyoung Huh, Brian Cheung, Tongzhou Wang, and Phillip Isola. 2024.
\newblock The platonic representation hypothesis.
\newblock In \emph{Proceedings of the 41st International Conference on Machine Learning}, Proceedings of Machine Learning Research. PMLR.

\bibitem[{Huo and Dunbar(2025)}]{huo2025ite}
Robin Huo and Ewan Dunbar. 2025.
\newblock Iterative refinement, not training objective, makes hubert behave differently from wav2vec 2.0.
\newblock In \emph{Proc. Interspeech 2025}, pages 261--265.

\bibitem[{Jia et~al.(2021)Jia, Yang, Xia, Chen, Parekh, Pham, Le, Sung, Li, and Duerig}]{align}
Chao Jia, Yinfei Yang, Ye~Xia, Yi-Ting Chen, Zarana Parekh, Hieu Pham, Quoc Le, Yun-Hsuan Sung, Zhen Li, and Tom Duerig. 2021.
\newblock Scaling up visual and vision-language representation learning with noisy text supervision.
\newblock In \emph{International conference on machine learning}, pages 4904--4916. PMLR.

\bibitem[{Jiang et~al.(2025)Jiang, Zhou, and Zhu}]{jiang2025tracing}
Jiachen Jiang, Jinxin Zhou, and Zhihui Zhu. 2025.
\newblock Tracing representation progression: Analyzing and enhancing layer-wise similarity.
\newblock In \emph{The Thirteenth International Conference on Learning Representations}.

\bibitem[{Kaplan et~al.(2020)Kaplan, McCandlish, Henighan, Brown, Chess, Child, Gray, Radford, Wu, and Amodei}]{kaplan2020scaling}
Jared Kaplan, Sam McCandlish, Tom Henighan, Tom~B Brown, Benjamin Chess, Rewon Child, Scott Gray, Alec Radford, Jeffrey Wu, and Dario Amodei. 2020.
\newblock Scaling laws for neural language models.
\newblock \emph{arXiv preprint arXiv:2001.08361}.

\bibitem[{Khan et~al.(2025)Khan, Leem, See, Wong, Zhang, and Fang}]{khan2025comprehensive}
Wasif Khan, Seowung Leem, Kyle~B See, Joshua~K Wong, Shaoting Zhang, and Ruogu Fang. 2025.
\newblock A comprehensive survey of foundation models in medicine.
\newblock \emph{IEEE Reviews in Biomedical Engineering}.

\bibitem[{Khosla et~al.(2024)Khosla, Williams, McDermott, and Kanwisher}]{khosla2024privileged}
Meenakshi Khosla, Alex~H Williams, Josh McDermott, and Nancy Kanwisher. 2024.
\newblock Privileged representational axes in biological and artificial neural networks.
\newblock \emph{bioRxiv}, pages 2024--06.

\bibitem[{Kirillov et~al.(2023)Kirillov, Mintun, Ravi, Mao, Rolland, Gustafson, Xiao, Whitehead, Berg, Lo et~al.}]{kirillov2023segment}
Alexander Kirillov, Eric Mintun, Nikhila Ravi, Hanzi Mao, Chloe Rolland, Laura Gustafson, Tete Xiao, Spencer Whitehead, Alexander~C Berg, Wan-Yen Lo, and 1 others. 2023.
\newblock Segment anything.
\newblock In \emph{Proceedings of the IEEE/CVF international conference on computer vision}, pages 4015--4026.

\bibitem[{Klabunde et~al.(2025)Klabunde, Schumacher, Strohmaier, and Lemmerich}]{klabunde2025similarity}
Max Klabunde, Tobias Schumacher, Markus Strohmaier, and Florian Lemmerich. 2025.
\newblock Similarity of neural network models: A survey of functional and representational measures.
\newblock \emph{ACM Computing Surveys}, 57(9):1--52.

\bibitem[{Koh et~al.(2023)Koh, Salakhutdinov, and Fried}]{LtoI}
Jing~Yu Koh, Ruslan Salakhutdinov, and Daniel Fried. 2023.
\newblock Grounding language models to images for multimodal inputs and outputs.
\newblock In \emph{International Conference on Machine Learning}, pages 17283--17300. PMLR.

\bibitem[{Kornblith et~al.(2019)Kornblith, Norouzi, Lee, and Hinton}]{kornblith2019similarity}
Simon Kornblith, Mohammad Norouzi, Honglak Lee, and Geoffrey Hinton. 2019.
\newblock Similarity of neural network representations revisited.
\newblock In \emph{International conference on machine learning}, pages 3519--3529. PMLR.

\bibitem[{Lan et~al.(2024)Lan, Torr, Meek, Khakzar, Krueger, and Barez}]{lan2024quantifying}
Michael Lan, Philip Torr, Austin Meek, Ashkan Khakzar, David Krueger, and Fazl Barez. 2024.
\newblock Quantifying feature space universality across large language models via sparse autoencoders.
\newblock \emph{arXiv preprint arXiv:2410.06981}.

\bibitem[{Lan et~al.(2020)Lan, Chen, Goodman, Gimpel, Sharma, and Soricut}]{Lan2020ALBERT}
Zhenzhong Lan, Mingda Chen, Sebastian Goodman, Kevin Gimpel, Piyush Sharma, and Radu Soricut. 2020.
\newblock Albert: A lite bert for self-supervised learning of language representations.
\newblock In \emph{International Conference on Learning Representations}.

\bibitem[{Lee et~al.(2025)Lee, Weber, Vi{\'e}gas, and Wattenberg}]{lee2025shared}
Andrew Lee, Melanie Weber, Fernanda Vi{\'e}gas, and Martin Wattenberg. 2025.
\newblock Shared global and local geometry of language model embeddings.
\newblock In \emph{Second Conference on Language Modeling}.

\bibitem[{Lee et~al.(2024)Lee, Liu, Sinhamahapatra, and Niehues}]{lee2024multimodal}
Hyunji Lee, Danni Liu, Supriti Sinhamahapatra, and Jan Niehues. 2024.
\newblock How do multimodal foundation models encode text and speech? an analysis of cross-lingual and cross-modal representations.
\newblock \emph{arXiv preprint arXiv:2411.17666}.

\bibitem[{Lenc and Vedaldi(2015)}]{lenc2015understanding}
Karel Lenc and Andrea Vedaldi. 2015.
\newblock Understanding image representations by measuring their equivariance and equivalence.
\newblock In \emph{Proceedings of the IEEE conference on computer vision and pattern recognition}, pages 991--999.

\bibitem[{Li et~al.(2022)Li, Li, Xiong, and Hoi}]{blip}
Junnan Li, Dongxu Li, Caiming Xiong, and Steven Hoi. 2022.
\newblock Blip: Bootstrapping language-image pre-training for unified vision-language understanding and generation.
\newblock In \emph{International conference on machine learning}, pages 12888--12900. PMLR.

\bibitem[{Li(2025)}]{li2025review}
Xinzhe Li. 2025.
\newblock A review of prominent paradigms for {LLM}-based agents: Tool use, planning (including {RAG}), and feedback learning.
\newblock In \emph{Proceedings of the 31st International Conference on Computational Linguistics}, pages 9760--9779.

\bibitem[{Li et~al.(2015)Li, Yosinski, Clune, Lipson, and Hopcroft}]{li2015convergent}
Yixuan Li, Jason Yosinski, Jeff Clune, Hod Lipson, and John Hopcroft. 2015.
\newblock Convergent learning: Do different neural networks learn the same representations?
\newblock \emph{arXiv preprint arXiv:1511.07543}.

\bibitem[{Li et~al.(2025)Li, Li, Shi, Farimani, Kluger, Yang, and Wang}]{li2025dual}
Zijie Li, Henry Li, Yichun Shi, Amir~Barati Farimani, Yuval Kluger, Linjie Yang, and Peng Wang. 2025.
\newblock Dual diffusion for unified image generation and understanding.
\newblock In \emph{Proceedings of the Computer Vision and Pattern Recognition Conference}, pages 2779--2790.

\bibitem[{Liu et~al.(2023)Liu, Li, Wu, and Lee}]{llava}
Haotian Liu, Chunyuan Li, Qingyang Wu, and Yong~Jae Lee. 2023.
\newblock Visual instruction tuning.
\newblock \emph{Advances in neural information processing systems}, 36:34892--34916.

\bibitem[{Liu et~al.(2019)Liu, Ott, Goyal, Du, Joshi, Chen, Levy, Lewis, Zettlemoyer, and Stoyanov}]{roberta}
Yinhan Liu, Myle Ott, Naman Goyal, Jingfei Du, Mandar Joshi, Danqi Chen, Omer Levy, Mike Lewis, Luke Zettlemoyer, and Veselin Stoyanov. 2019.
\newblock Roberta: A robustly optimized bert pretraining approach.
\newblock \emph{arXiv preprint arXiv:1907.11692}.

\bibitem[{Liu et~al.(2022)Liu, Mao, Wu, Feichtenhofer, Darrell, and Xie}]{convnet}
Zhuang Liu, Hanzi Mao, Chao-Yuan Wu, Christoph Feichtenhofer, Trevor Darrell, and Saining Xie. 2022.
\newblock A convnet for the 2020s.
\newblock \emph{Proceedings of the IEEE/CVF Conference on Computer Vision and Pattern Recognition (CVPR)}.

\bibitem[{Lu et~al.(2025)Lu, Wang, Yang, Zhang, Jenni, and Fu}]{lu2025indra}
Jianglin Lu, Hailing Wang, Kuo Yang, Yitian Zhang, Simon Jenni, and Yun Fu. 2025.
\newblock The indra representation hypothesis.
\newblock \emph{Advances in Neural Information Processing Systems}.

\bibitem[{Maniparambil et~al.(2024)Maniparambil, Akshulakov, Djilali, El~Amine~Seddik, Narayan, Mangalam, and O'Connor}]{Maniparambil}
Mayug Maniparambil, Raiymbek Akshulakov, Yasser Abdelaziz~Dahou Djilali, Mohamed El~Amine~Seddik, Sanath Narayan, Karttikeya Mangalam, and Noel~E. O'Connor. 2024.
\newblock Do vision and language encoders represent the world similarly?
\newblock In \emph{Proceedings of the IEEE/CVF Conference on Computer Vision and Pattern Recognition (CVPR)}.

\bibitem[{Merullo et~al.(2023)Merullo, Castricato, Eickhoff, and Pavlick}]{linearly}
Jack Merullo, Louis Castricato, Carsten Eickhoff, and Ellie Pavlick. 2023.
\newblock Linearly mapping from image to text space.
\newblock In \emph{The Eleventh International Conference on Learning Representations}.

\bibitem[{Mohamed et~al.(2022)Mohamed, Lee, Borgholt, Havtorn, Edin, Igel, Kirchhoff, Li, Livescu, Maal{\o}e et~al.}]{mohamed2022self}
Abdelrahman Mohamed, Hung-yi Lee, Lasse Borgholt, Jakob~D Havtorn, Joakim Edin, Christian Igel, Katrin Kirchhoff, Shang-Wen Li, Karen Livescu, Lars Maal{\o}e, and 1 others. 2022.
\newblock Self-supervised speech representation learning: A review.
\newblock \emph{IEEE Journal of Selected Topics in Signal Processing}, 16(6):1179--1210.

\bibitem[{Moor et~al.(2023)Moor, Banerjee, Abad, Krumholz, Leskovec, Topol, and Rajpurkar}]{moor2023foundation}
Michael Moor, Oishi Banerjee, Zahra Shakeri~Hossein Abad, Harlan~M Krumholz, Jure Leskovec, Eric~J Topol, and Pranav Rajpurkar. 2023.
\newblock Foundation models for generalist medical artificial intelligence.
\newblock \emph{Nature}, 616(7956):259--265.

\bibitem[{Morcos et~al.(2018)Morcos, Raghu, and Bengio}]{morcos2018insights}
Ari Morcos, Maithra Raghu, and Samy Bengio. 2018.
\newblock Insights on representational similarity in neural networks with canonical correlation.
\newblock \emph{Advances in neural information processing systems}, 31.

\bibitem[{Moschella et~al.(2023)Moschella, Maiorca, Fumero, Norelli, Locatello, and Rodol{\`a}}]{moschella2022relative}
Luca Moschella, Valentino Maiorca, Marco Fumero, Antonio Norelli, Francesco Locatello, and Emanuele Rodol{\`a}. 2023.
\newblock Relative representations enable zero-shot latent space communication.
\newblock In \emph{The Eleventh International Conference on Learning Representations}.

\bibitem[{Ng et~al.(2023)Ng, Subramanian, Klein, Kanazawa, Darrell, and Ginosar}]{ng2023can}
Evonne Ng, Sanjay Subramanian, Dan Klein, Angjoo Kanazawa, Trevor Darrell, and Shiry Ginosar. 2023.
\newblock Can language models learn to listen?
\newblock In \emph{Proceedings of the IEEE/CVF International Conference on Computer Vision}, pages 10083--10093.

\bibitem[{Ngo and Kim(2024)}]{ngo2024language}
Jerry Ngo and Yoon Kim. 2024.
\newblock What do language models hear? probing for auditory representations in language models.
\newblock In \emph{Proceedings of the 62nd Annual Meeting of the Association for Computational Linguistics}, pages 5435--5448.

\bibitem[{Nguyen et~al.(2021)Nguyen, Raghu, and Kornblith}]{nguyen2021do}
Thao Nguyen, Maithra Raghu, and Simon Kornblith. 2021.
\newblock Do wide and deep networks learn the same things? uncovering how neural network representations vary with width and depth.
\newblock In \emph{International Conference on Learning Representations}.

\bibitem[{Nguyen et~al.(2024)Nguyen, Wallingford, Santy, Ma, Oh, Schmidt, Koh, and Krishna}]{nguyen2024multilingual}
Thao Nguyen, Matthew Wallingford, Sebastin Santy, Wei-Chiu Ma, Sewoong Oh, Ludwig Schmidt, Pang Wei~W Koh, and Ranjay Krishna. 2024.
\newblock Multilingual diversity improves vision-language representations.
\newblock \emph{Advances in Neural Information Processing Systems}, 37:91430--91459.

\bibitem[{Ollerenshaw et~al.(2021)Ollerenshaw, Jalal, and Hain}]{ollerenshaw21}
Anna Ollerenshaw, Md~Asif Jalal, and Thomas Hain. 2021.
\newblock Insights on neural representations for end-to-end speech recognition.
\newblock In \emph{Interspeech}.

\bibitem[{Oozeer et~al.(2025)Oozeer, Nathawani, Prakash, Lan, HARRASSE, and Abdullah}]{oozeer2025}
Narmeen~Fatimah Oozeer, Dhruv Nathawani, Nirmalendu Prakash, Michael Lan, Abir HARRASSE, and Amir Abdullah. 2025.
\newblock Activation space interventions can be transferred between large language models.
\newblock In \emph{Forty-second International Conference on Machine Learning}.

\bibitem[{Oquab et~al.(2023)Oquab, Darcet, Moutakanni, Vo, Szafraniec, Khalidov, Fernandez, Haziza, Massa, El-Nouby et~al.}]{dinov2}
Maxime Oquab, Timoth{\'e}e Darcet, Th{\'e}o Moutakanni, Huy Vo, Marc Szafraniec, Vasil Khalidov, Pierre Fernandez, Daniel Haziza, Francisco Massa, Alaaeldin El-Nouby, and 1 others. 2023.
\newblock Dinov2: Learning robust visual features without supervision.
\newblock \emph{arXiv preprint arXiv:2304.07193}.

\bibitem[{Park et~al.(2024)Park, Choe, and Veitch}]{park2024linear}
Kiho Park, Yo~Joong Choe, and Victor Veitch. 2024.
\newblock The linear representation hypothesis and the geometry of large language models.
\newblock In \emph{International Conference on Machine Learning}, pages 39643--39666. PMLR.

\bibitem[{Pasad et~al.(2024)Pasad, Chien, Settle, and Livescu}]{pasad2024self}
Ankita Pasad, Chung-Ming Chien, Shane Settle, and Karen Livescu. 2024.
\newblock What do self-supervised speech models know about words?
\newblock \emph{Transactions of the Association for Computational Linguistics}, 12:372--391.

\bibitem[{Pasad et~al.(2023)Pasad, Shi, and Livescu}]{pasad2023co}
Ankita Pasad, Bowen Shi, and Karen Livescu. 2023.
\newblock Comparative layer-wise analysis of self-supervised speech models.
\newblock In \emph{ICASSP 2023-2023 IEEE International Conference on Acoustics, Speech and Signal Processing (ICASSP)}, pages 1--5. IEEE.

\bibitem[{Pham et~al.(2023)Pham, Matsui, and Chikazoe}]{biology12101330}
Trung~Quang Pham, Teppei Matsui, and Junichi Chikazoe. 2023.
\newblock Evaluation of the hierarchical correspondence between the human brain and artificial neural networks: A review.
\newblock \emph{Biology}, 12(10).

\bibitem[{Phang et~al.(2021)Phang, Liu, and Bowman}]{phang2021fine}
Jason Phang, Haokun Liu, and Samuel~R Bowman. 2021.
\newblock Fine-tuned transformers show clusters of similar representations across layers.
\newblock \emph{arXiv preprint arXiv:2109.08406}.

\bibitem[{Radford et~al.(2021)Radford, Kim, Hallacy, Ramesh, Goh, Agarwal, Sastry, Askell, Mishkin, Clark et~al.}]{clip}
Alec Radford, Jong~Wook Kim, Chris Hallacy, Aditya Ramesh, Gabriel Goh, Sandhini Agarwal, Girish Sastry, Amanda Askell, Pamela Mishkin, Jack Clark, and 1 others. 2021.
\newblock Learning transferable visual models from natural language supervision.
\newblock In \emph{International conference on machine learning}, pages 8748--8763. PMLR.

\bibitem[{Radford et~al.(2023)Radford, Kim, Xu, Brockman, McLeavey, and Sutskever}]{whisper}
Alec Radford, Jong~Wook Kim, Tao Xu, Greg Brockman, Christine McLeavey, and Ilya Sutskever. 2023.
\newblock Robust speech recognition via large-scale weak supervision.
\newblock In \emph{International conference on machine learning}, pages 28492--28518. PMLR.

\bibitem[{Radford et~al.(2019)Radford, Wu, Child, Luan, Amodei, Sutskever et~al.}]{gpt2}
Alec Radford, Jeffrey Wu, Rewon Child, David Luan, Dario Amodei, Ilya Sutskever, and 1 others. 2019.
\newblock Language models are unsupervised multitask learners.
\newblock \emph{OpenAI blog}, 1(8):9.

\bibitem[{Raffel et~al.(2020)Raffel, Shazeer, Roberts, Lee, Narang, Matena, Zhou, Li, and Liu}]{T5}
Colin Raffel, Noam Shazeer, Adam Roberts, Katherine Lee, Sharan Narang, Michael Matena, Yanqi Zhou, Wei Li, and Peter~J Liu. 2020.
\newblock Exploring the limits of transfer learning with a unified text-to-text transformer.
\newblock \emph{Journal of machine learning research}, 21(140):1--67.

\bibitem[{Raghu et~al.(2017{\natexlab{a}})Raghu, Gilmer, Yosinski, and Sohl-Dickstein}]{raghu2017svcca}
Maithra Raghu, Justin Gilmer, Jason Yosinski, and Jascha Sohl-Dickstein. 2017{\natexlab{a}}.
\newblock Svcca: Singular vector canonical correlation analysis for deep learning dynamics and interpretability.
\newblock \emph{Advances in neural information processing systems}, 30.

\bibitem[{Raghu et~al.(2017{\natexlab{b}})Raghu, Gilmer, Yosinski, and Sohl-Dickstein}]{Raghu20175}
Maithra Raghu, Justin Gilmer, Jason Yosinski, and Jascha Sohl-Dickstein. 2017{\natexlab{b}}.
\newblock Svcca: Singular vector canonical correlation analysis for deep learning dynamics and interpretability.
\newblock In \emph{Advances in Neural Information Processing Systems}, volume~30.

\bibitem[{Raghu et~al.(2021)Raghu, Unterthiner, Kornblith, Zhang, and Dosovitskiy}]{raghu2021vision}
Maithra Raghu, Thomas Unterthiner, Simon Kornblith, Chiyuan Zhang, and Alexey Dosovitskiy. 2021.
\newblock Do vision transformers see like convolutional neural networks?
\newblock \emph{Advances in neural information processing systems}, 34:12116--12128.

\bibitem[{Rajendran et~al.(2024)Rajendran, Buchholz, Aragam, Sch{\"o}lkopf, and Ravikumar}]{rajendran2024causal}
Goutham Rajendran, Simon Buchholz, Bryon Aragam, Bernhard Sch{\"o}lkopf, and Pradeep Ravikumar. 2024.
\newblock From causal to concept-based representation learning.
\newblock \emph{Advances in Neural Information Processing Systems}, 37:101250--101296.

\bibitem[{Raugel et~al.(2025)Raugel, Szafraniec, Vo, Couprie, Labatut, Bojanowski, Wyart, and King}]{raugel2025disentangling}
Jos{\'e}phine Raugel, Marc Szafraniec, Huy~V Vo, Camille Couprie, Patrick Labatut, Piotr Bojanowski, Valentin Wyart, and Jean-R{\'e}mi King. 2025.
\newblock Disentangling the factors of convergence between brains and computer vision models.
\newblock \emph{arXiv preprint arXiv:2508.18226}.

\bibitem[{Ravi et~al.(2025)Ravi, Gabeur, Hu, Hu, Ryali, Ma, Khedr, R{\"a}dle, Rolland, Gustafson, Mintun, Pan, Alwala, Carion, Wu, Girshick, Dollar, and Feichtenhofer}]{ravi2025sam}
Nikhila Ravi, Valentin Gabeur, Yuan-Ting Hu, Ronghang Hu, Chaitanya Ryali, Tengyu Ma, Haitham Khedr, Roman R{\"a}dle, Chloe Rolland, Laura Gustafson, Eric Mintun, Junting Pan, Kalyan~Vasudev Alwala, Nicolas Carion, Chao-Yuan Wu, Ross Girshick, Piotr Dollar, and Christoph Feichtenhofer. 2025.
\newblock {SAM} 2: Segment anything in images and videos.
\newblock In \emph{The Thirteenth International Conference on Learning Representations}.

\bibitem[{Riera et~al.(2023)Riera, Cerdeiro, Pepino, and Ferrer}]{riera2023phone}
Pablo Riera, Manuela Cerdeiro, Leonardo Pepino, and Luciana Ferrer. 2023.
\newblock Phone and speaker spatial organization in self-supervised speech representations.
\newblock In \emph{2023 IEEE International Conference on Acoustics, Speech, and Signal Processing Workshops (ICASSPW)}, pages 1--5. IEEE.

\bibitem[{Rinaldi et~al.(2025)Rinaldi, Capitani, Bonicelli, Crisostomi, Bolelli, FICARRA, Rodol{\`a}, Calderara, and Porrello}]{rinaldi2025update}
Filippo Rinaldi, Giacomo Capitani, Lorenzo Bonicelli, Donato Crisostomi, Federico Bolelli, ELISA FICARRA, Emanuele Rodol{\`a}, Simone Calderara, and Angelo Porrello. 2025.
\newblock Update your transformer to the latest release: Re-basin of task vectors.
\newblock In \emph{Forty-second International Conference on Machine Learning}.

\bibitem[{Roeder et~al.(2021)Roeder, Metz, and Kingma}]{roeder2021linear}
Geoffrey Roeder, Luke Metz, and Durk Kingma. 2021.
\newblock On linear identifiability of learned representations.
\newblock In \emph{International Conference on Machine Learning}, pages 9030--9039. PMLR.

\bibitem[{Sanh et~al.(2022)Sanh, Webson, Raffel, Bach, Sutawika, Alyafeai, Chaffin, Stiegler, Raja, Dey, Bari, Xu, Thakker, Sharma, Szczechla, Kim, Chhablani, Nayak, Datta, Chang, Jiang, Wang, Manica, Shen, Yong, Pandey, Bawden, Wang, Neeraj, Rozen, Sharma, Santilli, Fevry, Fries, Teehan, Scao, Biderman, Gao, Wolf, and Rush}]{sanh2022multitask}
Victor Sanh, Albert Webson, Colin Raffel, Stephen Bach, Lintang Sutawika, Zaid Alyafeai, Antoine Chaffin, Arnaud Stiegler, Arun Raja, Manan Dey, M~Saiful Bari, Canwen Xu, Urmish Thakker, Shanya~Sharma Sharma, Eliza Szczechla, Taewoon Kim, Gunjan Chhablani, Nihal Nayak, Debajyoti Datta, and 21 others. 2022.
\newblock Multitask prompted training enables zero-shot task generalization.
\newblock In \emph{International Conference on Learning Representations}.

\bibitem[{Schneider et~al.(2019)Schneider, Baevski, Collobert, and Auli}]{wav2vec}
Steffen Schneider, Alexei Baevski, Ronan Collobert, and Michael Auli. 2019.
\newblock wav2vec: Unsupervised pre-training for speech recognition.
\newblock \emph{arXiv preprint arXiv:1904.05862}.

\bibitem[{Shahbazi et~al.(2021)Shahbazi, Shirali, Aghajan, and Nili}]{shahbazi2021using}
Mahdiyar Shahbazi, Ali Shirali, Hamid Aghajan, and Hamed Nili. 2021.
\newblock Using distance on the riemannian manifold to compare representations in brain and in models.
\newblock \emph{NeuroImage}, 239:118271.

\bibitem[{Sharma et~al.(2024)Sharma, Shaham, Baradad, Fu, Rodriguez-Munoz, Duggal, Isola, and Torralba}]{visioncheckup}
Pratyusha Sharma, Tamar~Rott Shaham, Manel Baradad, Stephanie Fu, Adrian Rodriguez-Munoz, Shivam Duggal, Phillip Isola, and Antonio Torralba. 2024.
\newblock A vision check-up for language models.
\newblock In \emph{Proceedings of the IEEE/CVF Conference on Computer Vision and Pattern Recognition}, pages 14410--14419.

\bibitem[{Sharon and Dar(2025)}]{sharon2025}
Yuval Sharon and Yehuda Dar. 2025.
\newblock How do the architecture and optimizer affect representation learning? on the training dynamics of representations in deep neural networks.
\newblock \emph{arXiv preprint arXiv:2405.17377}.

\bibitem[{Shekhar et~al.(2023)Shekhar, Bordes, Vincent, and Morcos}]{shekhar2023objectives}
Shashank Shekhar, Florian Bordes, Pascal Vincent, and Ari Morcos. 2023.
\newblock Objectives matter: Understanding the impact of self-supervised objectives on vision transformer representations.
\newblock \emph{arXiv preprint arXiv:2304.13089}.

\bibitem[{Siméoni et~al.(2025)Siméoni, Vo, Seitzer, Baldassarre, Oquab, Jose, Khalidov, Szafraniec, Yi, Ramamonjisoa, Massa, Haziza, Wehrstedt, Wang, Darcet, Moutakanni, Sentana, Roberts, Vedaldi, Tolan, Brandt, Couprie, Mairal, Jégou, Labatut, and Bojanowski}]{dinov3}
Oriane Siméoni, Huy~V. Vo, Maximilian Seitzer, Federico Baldassarre, Maxime Oquab, Cijo Jose, Vasil Khalidov, Marc Szafraniec, Seungeun Yi, Michaël Ramamonjisoa, Francisco Massa, Daniel Haziza, Luca Wehrstedt, Jianyuan Wang, Timothée Darcet, Théo Moutakanni, Leonel Sentana, Claire Roberts, Andrea Vedaldi, and 7 others. 2025.
\newblock Dinov3.

\bibitem[{Song et~al.(2007)Song, Smola, Gretton, Borgwardt, and Bedo}]{song2007supervised}
Le~Song, Alex Smola, Arthur Gretton, Karsten~M Borgwardt, and Justin Bedo. 2007.
\newblock Supervised feature selection via dependence estimation.
\newblock In \emph{Proceedings of the 24th international conference on Machine learning}, pages 823--830.

\bibitem[{Stoica et~al.(2024)Stoica, Bolya, Bjorner, Ramesh, Hearn, and Hoffman}]{stoica2024zipit}
G~Stoica, D~Bolya, J~Bjorner, P~Ramesh, T~Hearn, and J~Hoffman. 2024.
\newblock Zipit! merging models from different tasks without training.
\newblock In \emph{International Conference on Learning Representations}. International Conference on Learning Representations.

\bibitem[{Tan et~al.(2024)Tan, Chanin, Lynch, Paige, Kanoulas, Garriga-Alonso, and Kirk}]{tan2024analysing}
Daniel Tan, David Chanin, Aengus Lynch, Brooks Paige, Dimitrios Kanoulas, Adri{\`a} Garriga-Alonso, and Robert Kirk. 2024.
\newblock Analysing the generalisation and reliability of steering vectors.
\newblock \emph{Advances in Neural Information Processing Systems}, 37.

\bibitem[{Team et~al.(2023)Team, Anil, Borgeaud, Alayrac, Yu, Soricut, Schalkwyk, Dai, Hauth, Millican et~al.}]{team2023gemini}
Gemini Team, Rohan Anil, Sebastian Borgeaud, Jean-Baptiste Alayrac, Jiahui Yu, Radu Soricut, Johan Schalkwyk, Andrew~M Dai, Anja Hauth, Katie Millican, and 1 others. 2023.
\newblock Gemini: a family of highly capable multimodal models.
\newblock \emph{arXiv preprint arXiv:2312.11805}.

\bibitem[{Team et~al.(2025)Team, Abeyruwan, Ainslie, Alayrac, Arenas, Armstrong, Balakrishna, Baruch, Bauza, Blokzijl et~al.}]{team2025gemini}
Gemini~Robotics Team, Saminda Abeyruwan, Joshua Ainslie, Jean-Baptiste Alayrac, Montserrat~Gonzalez Arenas, Travis Armstrong, Ashwin Balakrishna, Robert Baruch, Maria Bauza, Michiel Blokzijl, and 1 others. 2025.
\newblock Gemini robotics: Bringing ai into the physical world.
\newblock \emph{arXiv preprint arXiv:2503.20020}.

\bibitem[{Touvron et~al.(2023)Touvron, Martin, Stone, Albert, Almahairi, Babaei, Bashlykov, Batra, Bhargava, Bhosale et~al.}]{touvron2023llama}
Hugo Touvron, Louis Martin, Kevin Stone, Peter Albert, Amjad Almahairi, Yasmine Babaei, Nikolay Bashlykov, Soumya Batra, Prajjwal Bhargava, Shruti Bhosale, and 1 others. 2023.
\newblock Llama 2: Open foundation and fine-tuned chat models.
\newblock \emph{arXiv preprint arXiv:2307.09288}.

\bibitem[{Vaswani et~al.(2017)Vaswani, Shazeer, Parmar, Uszkoreit, Jones, Gomez, Kaiser, and Polosukhin}]{NIPS2017_3f5ee243}
Ashish Vaswani, Noam Shazeer, Niki Parmar, Jakob Uszkoreit, Llion Jones, Aidan~N Gomez, \L~ukasz Kaiser, and Illia Polosukhin. 2017.
\newblock Attention is all you need.
\newblock In \emph{Advances in Neural Information Processing Systems}, volume~30.

\bibitem[{Waheed et~al.(2024)Waheed, Atwany, Raj, and Singh}]{waheed2024speech}
Abdul Waheed, Hanin Atwany, Bhiksha Raj, and Rita Singh. 2024.
\newblock What do speech foundation models not learn about speech?
\newblock \emph{arXiv preprint arXiv:2410.12948}.

\bibitem[{Wang et~al.(2020)Wang, Rao, Guo, Wang, Liu, and Guan}]{wang2020towards}
Chenxu Wang, Wei Rao, Wenna Guo, Pinghui Wang, Jun Liu, and Xiaohong Guan. 2020.
\newblock Towards understanding the instability of network embedding.
\newblock \emph{IEEE Transactions on Knowledge and Data Engineering}, 34(2):927--941.

\bibitem[{Wang et~al.(2025)Wang, Ge, Shu, Tang, Zhou, He, and Qiu}]{wang2025towards}
Junxuan Wang, Xuyang Ge, Wentao Shu, Qiong Tang, Yunhua Zhou, Zhengfu He, and Xipeng Qiu. 2025.
\newblock Towards universality: Studying mechanistic similarity across language model architectures.
\newblock In \emph{The Thirteenth International Conference on Learning Representations}.

\bibitem[{Wang et~al.(2018)Wang, Hu, Gu, Hu, Wu, He, and Hopcroft}]{wang2018towards}
Liwei Wang, Lunjia Hu, Jiayuan Gu, Zhiqiang Hu, Yue Wu, Kun He, and John Hopcroft. 2018.
\newblock Towards understanding learning representations: To what extent do different neural networks learn the same representation.
\newblock \emph{Advances in neural information processing systems}, 31.

\bibitem[{Wei et~al.(2022)Wei, Tay, Bommasani, Raffel, Zoph, Borgeaud, Yogatama, Bosma, Zhou, Metzler et~al.}]{wei2022emergent}
Jason Wei, Yi~Tay, Rishi Bommasani, Colin Raffel, Barret Zoph, Sebastian Borgeaud, Dani Yogatama, Maarten Bosma, Denny Zhou, Donald Metzler, and 1 others. 2022.
\newblock Emergent abilities of large language models.
\newblock \emph{arXiv preprint arXiv:2206.07682}.

\bibitem[{Wentworth(2021)}]{alignmentforum_natural_abstraction}
John Wentworth. 2021.
\newblock Testing the natural abstraction hypothesis.
\newblock \emph{AI Alignment Forum}.

\bibitem[{Woo et~al.(2023)Woo, Debnath, Hu, Chen, Liu, Kweon, and Xie}]{ConvNeXtV2}
Sanghyun Woo, Shoubhik Debnath, Ronghang Hu, Xinlei Chen, Zhuang Liu, In~So Kweon, and Saining Xie. 2023.
\newblock Convnext v2: Co-designing and scaling convnets with masked autoencoders.
\newblock In \emph{IEEE/CVF Conference on Computer Vision and Pattern Recognition (CVPR)}.

\bibitem[{Yang et~al.(2023)Yang, Zhang, Song, Hong, Xu, Zhao, Zhang, Cui, and Yang}]{yang2023diffusion}
Ling Yang, Zhilong Zhang, Yang Song, Shenda Hong, Runsheng Xu, Yue Zhao, Wentao Zhang, Bin Cui, and Ming-Hsuan Yang. 2023.
\newblock Diffusion models: A comprehensive survey of methods and applications.
\newblock \emph{ACM computing surveys}, 56(4):1--39.

\bibitem[{Yin et~al.(2024)Yin, Fu, Zhao, Li, Sun, Xu, and Chen}]{yin2024survey}
Shukang Yin, Chaoyou Fu, Sirui Zhao, Ke~Li, Xing Sun, Tong Xu, and Enhong Chen. 2024.
\newblock A survey on multimodal large language models.
\newblock \emph{National Science Review}, 11(12):nwae403.

\bibitem[{Yu et~al.(2022)Yu, Wang, Vasudevan, Yeung, Seyedhosseini, and Wu}]{coca}
Jiahui Yu, Zirui Wang, Vijay Vasudevan, Legg Yeung, Mojtaba Seyedhosseini, and Yonghui Wu. 2022.
\newblock Coca: Contrastive captioners are image-text foundation models.
\newblock \emph{arXiv preprint}.

\bibitem[{Yu et~al.(2025)Yu, Kwak, Jang, Jeong, Huang, Shin, and Xie}]{yu2025representation}
Sihyun Yu, Sangkyung Kwak, Huiwon Jang, Jongheon Jeong, Jonathan Huang, Jinwoo Shin, and Saining Xie. 2025.
\newblock Representation alignment for generation: Training diffusion transformers is easier than you think.
\newblock In \emph{The Thirteenth International Conference on Learning Representations}.

\bibitem[{Zhang et~al.(2024)Zhang, Wang, and Charton}]{zhang24instruction}
Dylan Zhang, Justin Wang, and Francois Charton. 2024.
\newblock Instruction diversity drives generalization to unseen tasks.
\newblock \emph{arXiv preprint arXiv:2402.10891}.

\bibitem[{Zhang et~al.(2023)Zhang, Herrmann, Hur, Cabrera, Jampani, Sun, and Yang}]{zhang2023a}
Junyi Zhang, Charles Herrmann, Junhwa Hur, Luisa~Polania Cabrera, Varun Jampani, Deqing Sun, and Ming-Hsuan Yang. 2023.
\newblock A tale of two features: Stable diffusion complements {DINO} for zero-shot semantic correspondence.
\newblock In \emph{Thirty-seventh Conference on Neural Information Processing Systems}.

\bibitem[{Zhang et~al.(2025)Zhang, Yang, and Agrawal}]{zhang2024assessing}
Le~Zhang, Qian Yang, and Aishwarya Agrawal. 2025.
\newblock Assessing and learning alignment of unimodal vision and language models.
\newblock In \emph{Proceedings of the IEEE/CVF Conference on Computer Vision and Pattern Recognition (CVPR)}, pages 14604--14614.

\bibitem[{Zhou et~al.(2024)Zhou, Li, Li, Yu, Liu, Wang, Zhang, Ji, Yan, He et~al.}]{zhou2024comprehensive}
Ce~Zhou, Qian Li, Chen Li, Jun Yu, Yixin Liu, Guangjing Wang, Kai Zhang, Cheng Ji, Qiben Yan, Lifang He, and 1 others. 2024.
\newblock A comprehensive survey on pretrained foundation models: A history from bert to chatgpt.
\newblock \emph{International Journal of Machine Learning and Cybernetics}, pages 1--65.

\bibitem[{Zhou et~al.(2025)Zhou, Pan, LeCun, and Pinto}]{zhou2025dinowm}
Gaoyue Zhou, Hengkai Pan, Yann LeCun, and Lerrel Pinto. 2025.
\newblock {DINO}-{WM}: World models on pre-trained visual features enable zero-shot planning.
\newblock In \emph{Forty-second International Conference on Machine Learning}.

\end{thebibliography}

\end{document}